\documentclass[10pt,twocolumn,letterpaper]{article}
\usepackage{iccv}
\usepackage{times}
\usepackage{epsfig}
\usepackage{graphicx}
\usepackage{amsmath}
\usepackage{amssymb}
\usepackage{subcaption}
\usepackage{adjustbox}
\usepackage{bbm}
\usepackage{mathrsfs}
\usepackage{amsmath,amssymb,amsfonts,amsthm}
\usepackage{enumitem}
\newlist{assumptionlist}{enumerate}{1}
\setlist[assumptionlist,1]{label={\textbf{A\arabic*:}},labelwidth={40pt},leftmargin=*}

\usepackage[breaklinks=true,bookmarks=false]{hyperref}

\iccvfinalcopy 

\def\iccvPaperID{****}

\begin{document}

\title{An Empirical Study of Uncertainty in Polygon Annotation and the Impact of Quality Assurance}

\author{
Eric Zimmermann\footnotemark[1]\quad Justin Szeto\footnotemark[1]\quad Frederic Ratle \\
Sama \\
Montréal, Québec \\
{\tt\small \{ezimmermann,jszeto,fratle\}@samasource.org}\\
}

\maketitle

\begin{abstract}
   Polygons are a common annotation format used for quickly annotating objects in instance segmentation tasks. However, many real-world annotation projects request near pixel-perfect labels. While strict pixel guidelines may appear to be the solution to a successful project, practitioners often fail to assess the feasibility of the work requested, and overlook common factors that may challenge the notion of quality. This paper aims to examine and quantify the inherent uncertainty for polygon annotations and the role that quality assurance plays in minimizing its effect. To this end, we conduct an analysis on multi-rater polygon annotations for several objects from the MS-COCO dataset. The results demonstrate that the reliability of a polygon annotation is dependent on a reviewing procedure, as well as the scene and shape complexity.
\end{abstract}

\section{Introduction}\footnotetext[1]{Accepted at ICCV 2023 DataComp Workshop. First two authors contributed equally.}
Over the past decade, the emergence of deep learning in computer vision has enabled applications to interpret and understand visual information with increasing accuracy. At the heart of this progress lies the crucial role of annotations in training, validating, and fine-tuning machine learning (ML) models through supervised learning. Annotations provide the necessary context and labelled data that empower algorithms to recognize and extract meaningful insights from images and videos.

In order to acquire annotations, human annotators meticulously label various visual elements and attributes of a scene. These annotations can be provided as class labels, semantic descriptors, bounding boxes, or dense contours described by masks or polygons. These annotations serve as ground truth references and are fundamental to the development and deployment of reliable ML systems. 

\begin{figure}[t]
    \centering
    \includegraphics[width=\columnwidth]{ 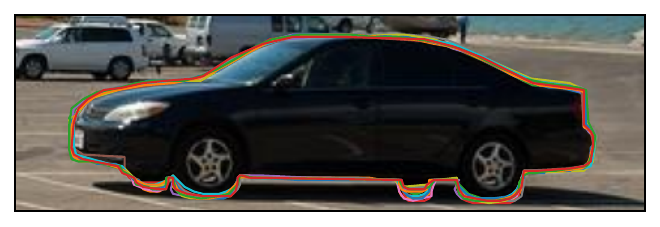} 
    \caption{Variability in annotating an object. The car tires are hidden by the car's shadow, thereby leading to large annotation uncertainty around the tires.}
    \label{fig:example_annotation_variability}
\end{figure}

A major challenge in computer vision annotations is the complexity and diversity of visual data. Images may contain a wide variety of dense and occluded objects at various resolutions and lighting conditions. These factors make it difficult to discern objects and lead to the notion of uncertainty, since it may not be possible to assign certain types of labels to ambiguous segments in an image (see example in Figure \ref{fig:example_annotation_variability}). Annotation quality relates to model quality, resulting in a need for perfection despite the task's overhead. The higher the quality of an annotation, the more challenging it may be to acquire the annotation. 

In a production setting, each annotation task is performed in accordance with instructional guidelines, and an emphasis is placed on fast completion time. In these workflows, a range of annotators with mixed levels of experience are required to complete a task with low or zero error tolerances using the polygon format. There is an expectation that quality can be met with guaranteed perfect pixel precision. However, these expectations fail to account for the constraints of the task, the ambiguity of the instructions, and the complexity of the scene. If a workflow is subject to a quality assurance (QA) review, these factors may result in additional reworks and wasted resources, which pose a risk to the project as a whole. It is possible to mitigate risk by better understanding the types of errors that are acceptable or unavoidable. This can be accomplished by synchronizing expectations across all parties involved in a project, by understanding the limits and uncertainties of a task, and by adapting reviews accordingly.

\section{Overview}
We study the presence of uncertainty and the effects of an additional quality assurance stage in a multi-rater annotation workflow. We summarize a group of polygons using the notion of a consensus shape and leverage it to describe global and local variability for sets of shapes. The choice of consensus model therefore dictates how reliable the downstream analysis may be. 

\subsection{Consensus}

 Let $s \subset \mathbb{R}^{2}$ be a closed shape with boundary contour $\partial{s}$. Given a set of $n$ shapes $S = \{s_i\}_{i=1}^n$, the asymmetric distance function $d(\partial{s_i}, \partial{s_j})$  measures the total surface distance between a pair of contours. For any point $p$ on a curve $\partial{s_i}$, we denote the corresponding point on $\partial{s_j}$ as $y_{\partial{s_j}}(p)$. The segment between $p$ and $\partial{s_j}$ is the geodesic \cite{shape_apx, integral_boundary_pde, medical_boundary_seg}. The asymmetric squared distance is defined as:
 
 \begin{equation}
     d(\partial{s_i}, \partial{s_j})^2 = \int_{\partial{s_i}} ||y_{\partial{s_j}}(p) - p||^2 dp.
 \end{equation}
 
For any set $S$ of curves, the mean shape $\bar{\mu}_s$ minimizes the asymmetric distance between all curves in $S$. Given the set $\Omega(\partial{s})$ of all curve boundaries, the mean curve $\bar{\mu}_s$ is defined as:

\begin{equation}
     \label{eqn:pde}
     \bar{\mu}_s = \operatorname*{argmin}_{\partial{\bar{\mu}_s} \in \Omega(\partial{s})} \sum_{s_i \in S} d(\partial{\bar{\mu}_s}, \partial{s_i})^2.
 \end{equation}
 
The mean curve is any curve that minimizes this partial differential equation, as per Eq \ref{eqn:pde}. A gradient flow can be used to iteratively construct an optimal curve, where the quality of the curve depends on the chosen initial conditions as well as the complexity and variability of the shapes in $S$ \cite{shape_apx}. In practice, we may find an adequate approximation to the mean curve in a discrete setting under a few strict assumptions. 
\begin{assumptionlist}
    \item The  set of all curves in $S$ all define the same underlying shape with minor variability.
    \item There exists an exact distance transform (EDT) \cite{exact_distance_transform} that approximates the  asymmetric distance over a subsection of the curve.
    \item The gradients of the distance function can be inferred from local measurements of the distance transform. 
\end{assumptionlist}
If all assumptions hold, we  solve for the mean curve by finding a contour that corresponds to the zero crossing of the Laplacian. The mean curve is computed using marching squares on the mean signed distance map based on the EDT for all contours in $S$. This representation is useful as it may be used to identify regions of interest with high disagreement between elements in $S$. Given an EDT map $D_{T_{s_i}}(p)$ at spatial location $p$,  the approximate lower bound to the average absolute difference between the boundaries of the mean contour and the set of shapes is computed as the average accumulation of distance over the mean contour. The expected boundary distance $d_B$ between the mean shape $\mu$ with contour length $|\partial{\mu}|$ and $S$ is therefore defined as:

\begin{equation}
    d_B(\mu, S) = \frac{1}{|\partial{\mu}| |S|} \sum_{s_i \in S} \int_{\partial{\mu}} D_{T_{s_i}}(p)dp.
\end{equation}

We note that the distance map approximates geodesics asymmetrically and is robust to large spikes in curvature. This distance therefore plays a role in the underestimation of the true distance measure.

\begin{figure}[h]
    \centering
    \hfill
     \begin{subfigure}[b]{0.24\columnwidth}
         \centering
         \includegraphics[width=\linewidth]{ 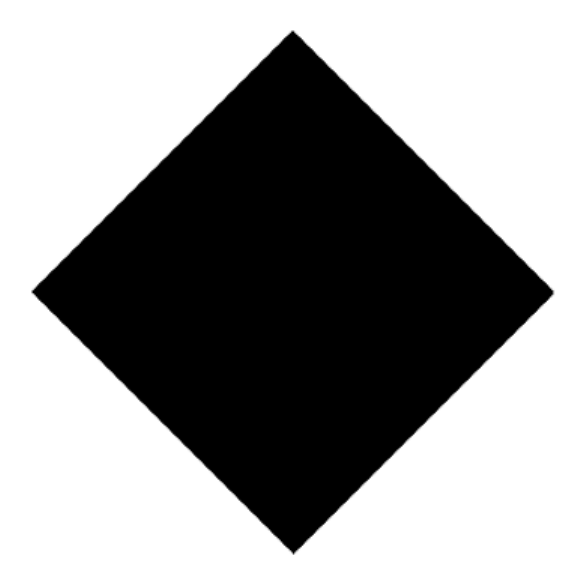}
         \caption{Diamond}
         \label{fig:diamond_polygon}
     \end{subfigure}
     \hfill
     \begin{subfigure}[b]{0.24\columnwidth}
         \includegraphics[width=\linewidth]{ 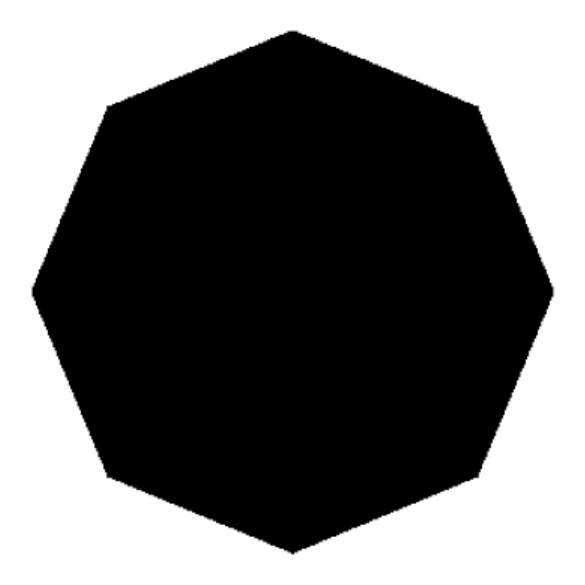}
         \caption{Octagon}
         \label{fig:octagon_polygon}
     \end{subfigure}
    \hfill 
     \begin{subfigure}[b]{0.24\columnwidth}
         \centering
         \includegraphics[width=\linewidth]{ 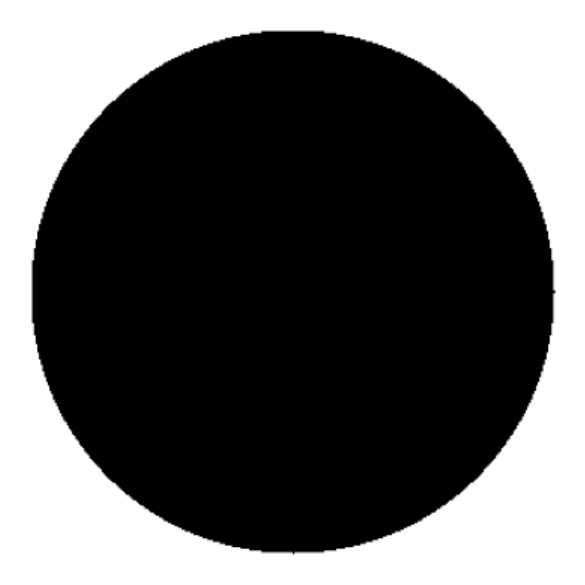}
         \caption{Circle}
         \label{fig:circle_polygon}
     \end{subfigure}
     \hfill
     \begin{subfigure}[b]{0.24\columnwidth}
         \centering
         \includegraphics[width=\linewidth]{ 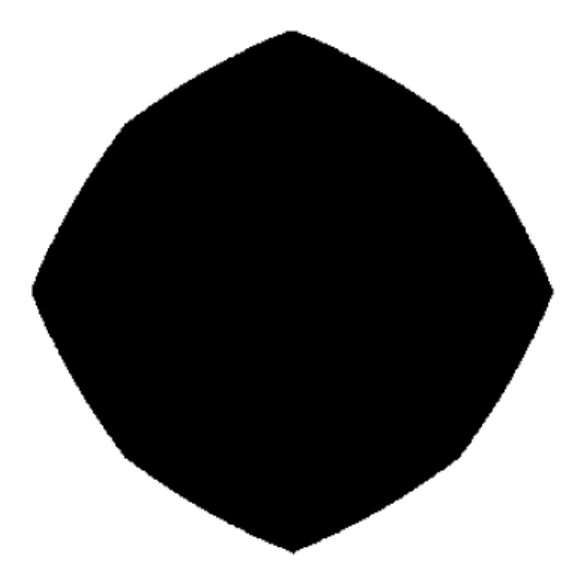}
         \caption{Consensus}
         \label{fig:mean_consensus_polygon}
     \end{subfigure}
    \caption{Example of a mean consensus polygon generated from a fair aggregation of a diamond, octagon, and circle.}
    \label{fig:example_augmentations}
\end{figure}

It is also possible to define the mode shape $\tilde{\mu}_s$ over the set of shapes $S$. The mode is defined at a point using a majority vote consensus over $\mathbb{R}^2$. The mode is not a particularly reliable consensus, as it may become jagged and discontinuous. It is also not clear how to then find corresponding distances between the mode curve and each curve that composes it. Moreover, it fails to capture distinct disagreement on subsections of a curve since it tends to converge to the majority trend. There are also other methods to compute consensus shapes that incorporate other priors and assumptions, such as Expectation-Maximization-based methods \cite{EM_consensus, STAPLE_consensus} and Reliability Aware Sequence Aggregation \cite{RASA_consensus}. Based on the aforementioned options, the mean curve is most suitable since it allows the user to fairly measure distances between the sets of curves with minimal bias.

\subsection{Rasterization}

Rasterization is the discretization process of mapping a polygon to a grid. It is used in the analysis of shapes and plays a role is the estimation of total uncertainty. Polygons are composed of a variable number of control vertices with linearly interpolated edges. Vertices are confident measurements of an object boundary since they are selected by an annotator, while edges contain subpixel decisions which are inherently prone to errors when discretized. Measurement uncertainty along edges can be minimized proportional to the resolution of the grid used to compute the consensus. Distance errors along each continuous edge are therefore minimized by trading off space for precision.

Consider a raster function $R_r: P \rightarrow \mathbb{R}^2$ that maps elements from a polygon $P$ to the spatial domain of its associated shape at a discrete resolution $r$. If the raster function operates in an all-touch setting, where any element on the grid is filled if a line segment in $P$ intersects the grid, the error between the area filled by the rasterization process tends towards the true area that $P$ encloses. Given area-measuring functions $A_P$ and $A_R$, $A_P$ is the count of filled tiles on the grid divided by the squared resolution and $A_P$ is the shoelace operation, it is clear that $\operatorname*{\lim}_{r \rightarrow \infty} A_R(R_r(P)) = A_P(P)$ geometrically.

\begin{figure}[h]
    \centering
    \includegraphics[width=\linewidth]{ 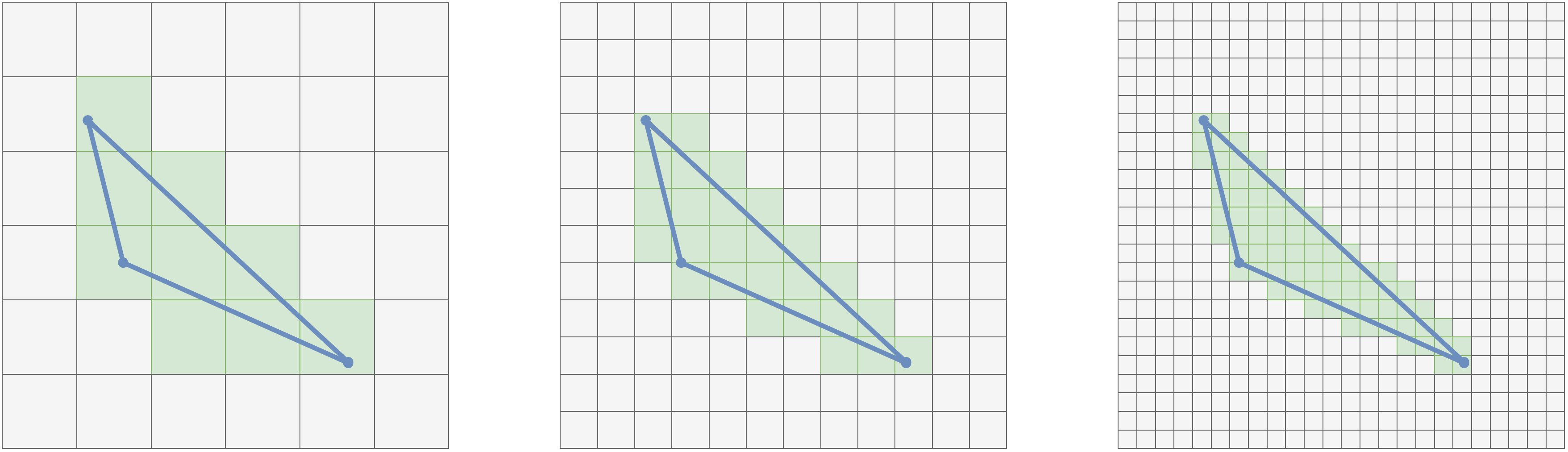}
    \caption{Rasterization process of an initial shape for increasingly dense grids. The area of the polygon converges to its true value as grid size increases.}
\end{figure}

\section{Dataset}
A set of images is acquired by manually selecting samples from MS-COCO \cite{coco}. Samples are selected based on a diversity criterion to ensure a mix of common classes composed of humans, cars, animals, and other miscellaneous common objects. A dataset sample is generated from a crop of an image based on a bounding box of interest. The crop is performed with a $10\%$ margin around the bounding box. Samples are selected to have no holes and minimal occluders in order to avoid inconsistencies in annotations due to scene ambiguities. In total, the dataset is composed of 24 samples that are each paired with a unique instruction set that states which areas should or should not be annotated. The instruction set is reviewed by a team of quality assurance specialists to ensure that instructions are consistent with standard project workflows. The annotators are tasked with labeling the contour of the object in question using a polygon tool, with zero pixel error tolerance. Annotators are allowed to perform zoom operations, but image enhancement tools are restricted. Once completed, each annotator receives feedback on their tasks by a dedicated quality assurance specialist and are tasked with repeating the process. Quality assurance is accomplished via a calibration session between the specialist and the annotators to establish a common understanding of the annotation precision needed and semantic errors to avoid. Note that the QA may bias the result towards a specific interpretation of the image and instructions, which is not always the same as ``the truth".

\section{Methodology and Results}
We analyze uncertainties aggregated across the dataset before and after the QA step and follow up with an analysis based on local curve segments with high variability. Analysis is done using the mean curve and an arc length parameterization is used to observe patterns on its subsections. 

\subsection{Aggregated Uncertainties}
\begin{figure}[h]
    \centering
    \includegraphics[width=\columnwidth]{ 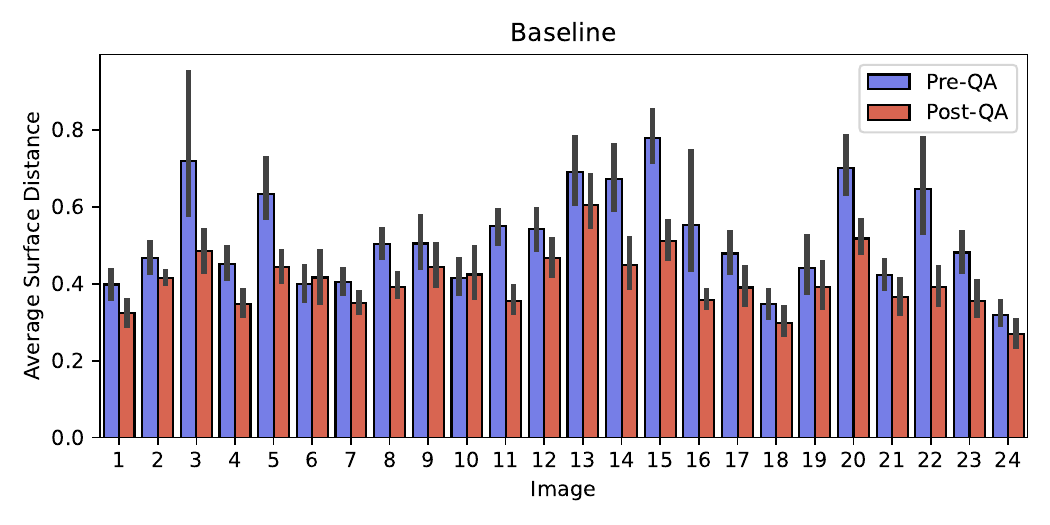} 
  \caption{Per-image mean distance and variance, and differences between pre- and post-QA. Error is dramatically reduced after a QA intervention.}
  \label{fig:baseline-pre-post}
\end{figure}

The mean shape is computed at a resolution of 8 and used to calculate the mean distance (measured in pixels) for each sample in the dataset. The process is repeated before and after a QA review. Average distances to the consensus are found to be $0.522\pm 0.650$ and $0.407\pm0.401$ respectively. A Welch's t-test at a $95\%$ confidence interval detects a statistically significant shift in the distribution of errors measured in pixels. The distributions of distances are presented in Figure \ref{fig:baseline-pre-post}.

\subsection{Local Contour Uncertainties}
\begin{figure}[h]
    \centering
    \includegraphics[width=\columnwidth]{ 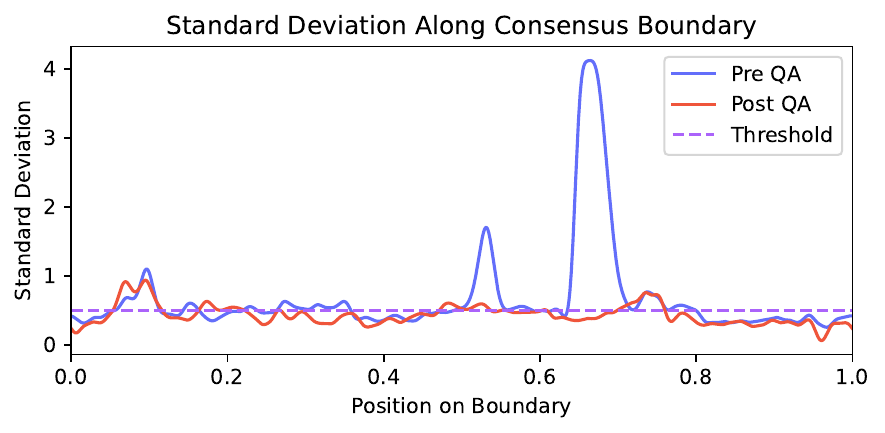}
    \caption{Contour variance over the arc length of the mean curve pre- and post-QA. The cutoff threshold is used to detect anomalous segments.}
    \label{fig:variance-plot}
\end{figure}

\begin{figure}[h]
     \begin{subfigure}[b]{0.48\linewidth}
         \centering
         \includegraphics[width=0.9\linewidth]{ 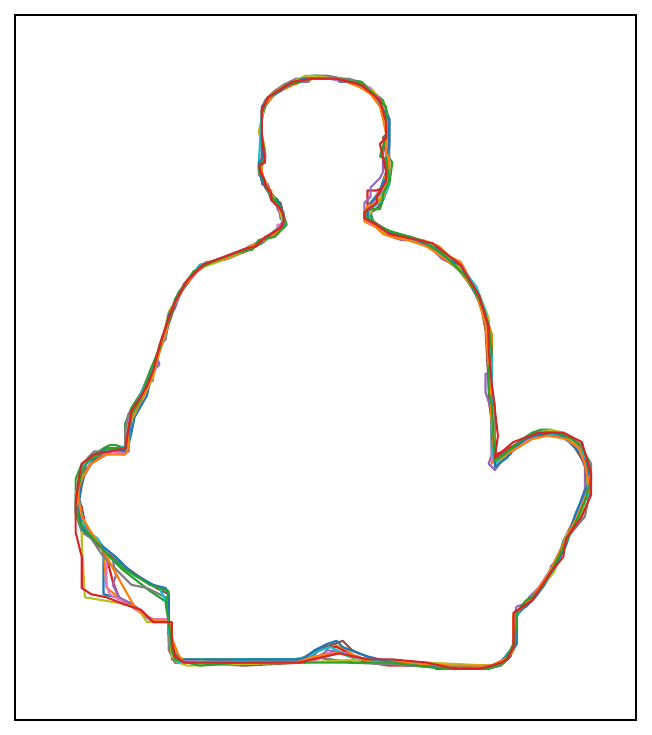}
         \caption{Pre-QA}
         \label{fig:pre-qa-polygon}
     \end{subfigure}
     \hfill
     \begin{subfigure}[b]{0.48\linewidth}
         \centering
         \includegraphics[width=0.9\linewidth]{ 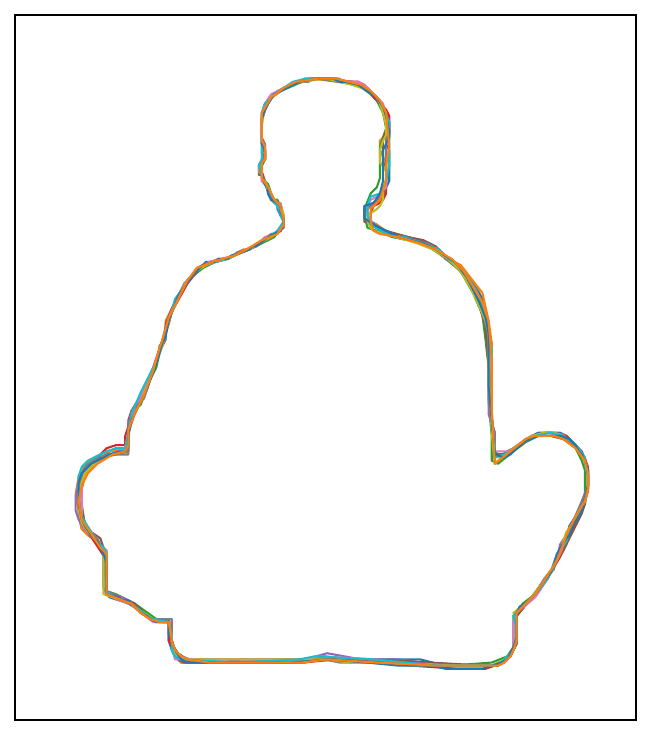}
         \caption{Post-QA}
         \label{fig:post-qa-polygon}
     \end{subfigure}
     \hfill
     \begin{subfigure}[b]{0.48\linewidth}
         \centering
         \includegraphics[width=0.9\linewidth]{ 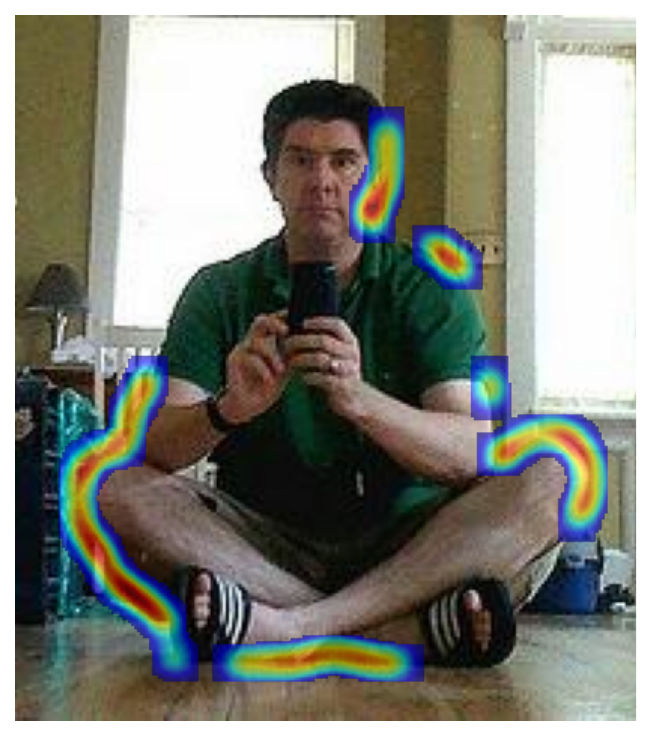}
         \caption{Pre-QA}
         \label{fig:pre-qa-heatmap}
     \end{subfigure}
     \hfill
     \begin{subfigure}[b]{0.48\linewidth}
         \centering
         \includegraphics[width=0.9\linewidth]{ 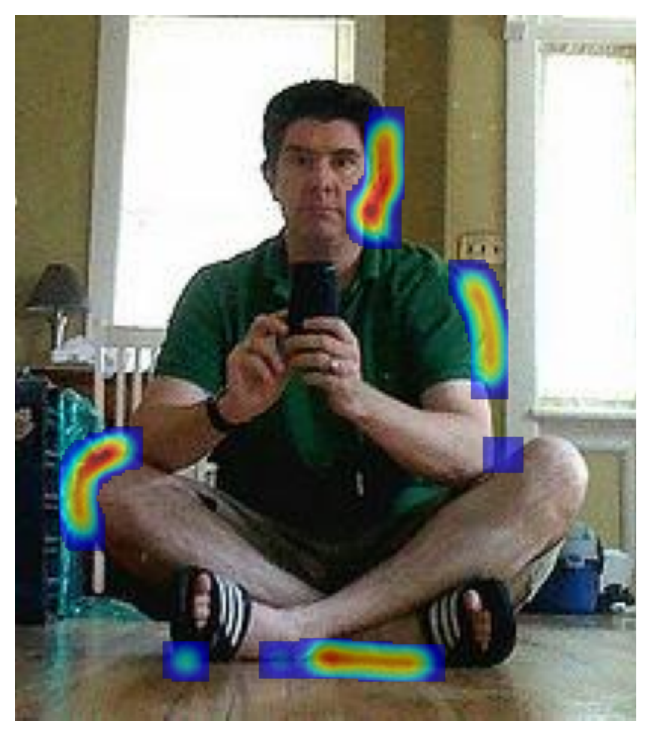}
         \caption{Post-QA}
         \label{fig:post-qa-heatmap}
     \end{subfigure}
        \caption{A set of polygons acquired pre- and post-QA. Heatmaps provide visual cues in regions with high local variance. Regardless of review, regions of low contrast and high curvature remain uncertain.}
        \label{fig:pre-post-poly-heat}
\end{figure}

Since important regions of interest cannot be flagged using aggregated statistics along the entirety of a curve, we quantify local variations between the curves and the consensus. Local analysis is performed over subsections of the curve using an arc length parameterization of the consensus for each image. The accumulation of the signed asymmetric distance over a subsection of the consensus is used to represent the standard deviation of a family of curves relative to its midpoint. A rejection threshold is used to find regions on the curve whose deviation exceeds this value. Results are visually verifiable as per Figure \ref{fig:variance-plot} and \ref{fig:pre-post-poly-heat} for a cutoff threshold of 0.5.

Figures \ref{fig:pre-post-poly-heat} demonstrates the impact of a QA step and how it leads to a correction. When observing the legs of the individual, it is unclear that there may be pants in the region where contrast is low. This is corrected by an executive decision based on style guidelines provided by a reviewer. On the other hand, regardless of a review, the variance along the head of the individual and below the legs could not be avoided. It is observed that these regions had poor contrast and higher relative curvature. It is noted that even though the post QA polygon looks reasonable, a rework may still be requested in a zero pixel tolerance setting due to minor disparities in the interpretations of the result. 

\subsection{Discussion}
Our analysis demonstrates that the QA process significantly enhanced style consistency and is a valuable step in the annotation process. Without an expert in the loop, stylistic variation can have a significant impact on the success of a project. The QA strategy is of utmost importance and cannot be understated. Rejections and reworks due to errors are the single largest factors limiting the success of a project. Small unavoidable issues may lead to flagging that greatly increase the total time spent working on a task and as a result, inflate the total cost of the venture. These costly interventions are associated with local variations. By understanding which errors can be tackled and which are inherent to an image, it is possible to reduce time, effort, and resources spent. 

Due to the limited data and cohort size, it is not possible to provide a statistical breakdown of what confounders have effects on local uncertainties. However, intuitively, there is a relationship between curvature, contrast, and uncertainty. Regions of low contrast and high curvature may lead to more uncertainty. An annotator in a known domain has an unspecifiable shape prior and additional context provided by the entire scene that cannot be estimated. These factors allow them to perform well in regions that would otherwise be difficult without this contextual information.

\section{Conclusion}
We demonstrate that a quality assurance review phase plays an important role in ensuring consistency throughout an annotation project, and that it is possible to assess the quality of a set of annotations using a mean consensus. Furthermore, we provide empirical results illustrating that the average spread of contours typically spans at least one pixel on a global level and often vastly exceeds acceptable ranges on a local level. Regions with inherently high uncertainty should not be simply reworked as they will lead to additional costs in a project and improvements cannot be guaranteed.

{\small
\bibliographystyle{ieee_fullname}
\bibliography{egbib}
}

\newpage
\appendix
\section{MS-COCO Image and Annotation IDs}\label{sec:coco-ids}
For reference purposes, the original MS-COCO image IDs and annotation IDs for all 24 samples in this study are provided in Table \ref{table:coco-ids}.

\begin{table}[h]
  \centering
  \begin{tabular}{|c|c|c|}
    \hline
    Image ID & COCO Image ID & COCO Annotation ID \\
    \hline
    \hline
    1 & 2592 & 678926 \\
    \hline
    2 & 82696 & 36233 \\
    \hline
    3 & 119677 & 1084188 \\
    \hline
    4 & 102805 & 351654 \\
    \hline
    5 & 96549 & 156368 \\
    \hline
    6 & 101420 & 1816501 \\
    \hline
    7 & 172595 & 109320 \\
    \hline
    8 & 197658 & 1706948 \\
    \hline
    9 & 216277 & 1048287 \\
    \hline
    10 & 232348 & 1128607 \\
    \hline
    11 & 5060 & 1731118 \\
    \hline
    12 & 181859 & 47665 \\
    \hline
    13 & 197022 & 1076210 \\
    \hline
    14 & 376284 & 346188 \\
    \hline
    15 & 376284 & 1388327 \\
    \hline
    16 & 416534 & 102024 \\
    \hline
    17 & 543043 & 135138 \\
    \hline
    18 & 8532 & 464797 \\
    \hline
    19 & 50679 & 1051538 \\
    \hline
    20 & 46804 & 63303 \\
    \hline
    21 & 46872 & 502460 \\
    \hline
    22 & 42070 & 164994 \\
    \hline
    23 & 52017 & 160713 \\
    \hline
    24 & 58539 & 559122 \\
    \hline
    \end{tabular}
  \caption{Mapping between image IDs used in the dataset and the original COCO image and annotation IDs}
  \label{table:coco-ids}
\end{table}

\end{document}